# DGL-RSIS: Decoupling Global Spatial Context and Local Class Semantics for Training-Free Remote Sensing Image Segmentation

Boyi Li [a], Ce Zhang [a,*], Richard M. Timmerman [a], Wenxuan Bao [b]

[a] *School of Geographical Sciences, University of Bristol, University Road, Bristol BS8 1SS, UK;*

[b] *Institute of Geographic Sciences and Natural Resources Research, Chinese Academy of Sciences, Beijing 100101, China*

* Corresponding author: Dr Ce Zhang (ce.zhang@bristol.ac.uk)

**Abstract**: The emergence of vision language models (VLMs) bridges the gap between vision and language, enabling multimodal understanding beyond traditional visual-only deep learning models. However, transferring VLMs from the natural image domain to remote sensing (RS) segmentation remains challenging due to the large domain gap and the diversity of RS inputs across tasks, particularly in open-vocabulary semantic segmentation (OVSS) and referring expression segmentation (RES). Here, we propose a training-free unified framework, termed DGL-RSIS, which decouples visual and textual representations and performs visual-language alignment at both local semantic and global contextual levels. Specifically, a Global–Local Decoupling (GLD) module decomposes textual inputs into local semantic tokens and global contextual tokens, while image inputs are partitioned into class-agnostic mask proposals. Then, a Local Visual-Textual Alignment (LVTA) module adaptively extracts context-aware visual features from the mask proposals and enriches textual features through knowledge-guided prompt engineering, achieving OVSS from a local perspective. Furthermore, a Global Visual–Textual Alignment (GVTA) module employs a global-enhanced Grad-CAM mechanism to capture contextual cues for referring expressions, followed by a mask selection module that integrates pixel-level activations into mask-level segmentation outputs, thereby achieving RES from a global perspective. Experiments on the iSAID (OVSS) and RRSIS-D (RES) benchmarks demonstrate that DGL-RSIS outperforms existing training-free approaches. Ablation studies further validate the effectiveness of each module. To the best of our knowledge, this is the first unified training-free framework for RS image segmentation, which effectively transfers the semantic capability of VLMs trained on natural images to the RS domain without additional training.

**Keywords**: Vision language model; Open-vocabulary semantic segmentation; Referring expression segmentation; Domain knowledge; Training-free

## 1. Introduction

Vision language models (VLMs) are designed to integrate visual and textual modalities for a more holistic understanding of scenes. By aligning visual content with linguistic semantics, VLMs are not only able to recognize objects in an image, but also to comprehend the semantic relationships between them. This enables impressive performance on vision-language tasks such as visual question answering (VQA) (Li et al., 2023), text-guided semantic segmentation (Ding et al., 2022) and visual grounding (Shen et al., 2024).

Recently, the use of VLMs in RS has gained increasing attention (X. Li et al., 2024). Compared to traditional vision-only models that rely heavily on supervised learning and often struggle with generalization, VLMs offer full potential as a robust alternative. By leveraging textual priors and semantic reasoning, VLMs exhibit strong zero-shot capabilities, especially when dealing with unseen or out-of-distribution classes—making them more suitable for real-world applications. With the growing availability of textual metadata associated with RS data, the community has begun to explore VLM-based solutions for RS tasks, particularly in image segmentation.

Two major segmentation paradigms have emerged under the VLM framework in RS image segmentation: open-vocabulary semantic segmentation (OVSS) (Li et al., 2025; J. Zhang et al., 2024) and referring expression segmentation (RES) (Liu et al., 2024a; Yuan et al., 2024). OVSS aims to segment RS images into semantic regions based on an open set of category names, including previously unseen classes. This task emphasizes fine-grained local class semantics. For example, Text2Seg (J. Zhang et al., 2024) introduced a text-guided segmentation framework for RS imagery by integrating several pretrained VLMs to facilitate semantic interpretation of objects in RS images. On the other hand, RES seeks to localize and segment specific objects described by detailed descriptions (e.g., "the small building on the left with a red roof"). This task not only requires understanding of local class semantics but also demands a grasp of global spatial context. Yuan et al. (2024) proposed the first large-scale RES benchmark (RRSIS) and a baseline model that incorporates shallow and deep features for multi-scale spatial understanding.

While VLMs are increasingly applied in RS, transferring them from the natural image domain to RS segmentation remains challenging. First, most VLMs are pretrained or fine-tuned on datasets such as COCO (Lin et al., 2014) or PhraseCut (Wu et al., 2020), which contain hundreds to thousands of fine-grained classes. In contrast, RS datasets often contain only a few coarse-grained categories (e.g., "water," "urban," "vegetation"), making supervised fine-tuning less effective and weakening generalization to unseen classes. Second, RS imagery is captured from a top-down perspective and includes objects of varying sizes due to differing spatial resolutions, introducing a significant domain gap between RS data and the natural images used in VLM pretraining.

Given these challenges, we aim to investigate effective strategies for transferring VLMs fine-tuned on natural images to RS segmentation tasks. Instead of relying on supervised adaptation, we explore a training-free strategy that directly exploits the strong vision-language alignment capabilities of VLMs to achieve universal RS segmentation, covering both OVSS and RES tasks within a unified framework. Motivated by the distinct requirements of these tasks—local semantics for OVSS and global context for RES—we propose a Global–Local Decoupling (GLD) module that decomposes vision–language inputs into local and global representations. To mitigate the domain gap between natural imagery and RS data, we design a Local Visual-Textual Alignment (LVTA) module for adaptively visual feature extraction and knowledge-guided textual feature enrichment. To jointly accommodate OVSS and RES, a Global Visual–Textual Alignment (GVTA) module is introduced, which employs a global-enhanced Grad-CAM mechanism to capture contextual cues for referring expressions and integrates pixel-level activations into mask-level segmentation outputs.

Our main contributions are summarized as follows. First, we propose a training-free framework for RS segmentation that decouples global context and local semantics, enabling unified support for both OVSS and RES tasks. Second, we improve local visual–textual alignment through a context-aware feature extraction mechanism and knowledge-guided prompt engineering, improving both visual discriminability and textual semantic understanding. Third, we design a global-enhanced Grad-CAM module for global visual–language alignment, which improves the model's ability to interpret spatial relationships in referring expressions and refines instance mask selection by combining Grad-CAM activations with mask proposals.

## 2. Related works

### 2.1 *Open-vocabulary semantic segmentation (OVSS)*

OVSS aims to segment an image into semantic regions defined by an open set of category names—potentially including new classes. Unlike traditional segmentation models constrained by closed-set annotations, OVSS leverages weak supervision signals, such as image-level labels or image–caption pairs, and more importantly, pretrained VLMs that align visual and textual modalities through contrastive learning. In typical OVSS pipelines, a VLM's text encoder embeds class names via template prompts (e.g., "a photo of a [*CLASS*]"), acting as frozen classifiers. These text features are matched with pixel- or region-level visual features to enable semantic segmentation.

Existing OVSS methods can be broadly categorized based on how segmentation masks are created. Pixel-based approaches, such as DenseCLIP (Rao et al., 2022), CLIP-S4 (He et al., 2023), and SegCLIP (Luo et al., 2023) formulate the task as a text–patch alignment problem. They compute similarity maps between image patches and class-specific textual embeddings using cosine similarity, treating the resulting heatmaps as semantic masks. These methods preserve spatial correspondence and produce fine-grained masks with minimal supervision. In contrast, mask-based approaches like GroupViT (Xu et al., 2022), MaskCLIP (Zhou et al., 2022), and ZegFormer (Ding et al., 2022) adopt a two-stage pipeline: generating class-agnostic region proposals, followed by assigning class labels through classification. This design offers flexibility in detecting multiple unseen categories but heavily relies on accurate region proposals.

In the RS domain, Chen and Bruzzone (2023) used a conditional U-Net model to predict semantic segmentation masks based on textual descriptions. They further leveraged the rich representations from a pretrained Contrastive Language–Image Pretraining (CLIP) model to align images and corresponding text embeddings through contrastive learning. SegCLIP (S. Zhang et al., 2024) extended CLIP for RS segmentation by incorporating prompt engineering and a cross-modal design. Metasegnet (Wang et al., 2024) introduces a cross-modal attention fusion module that integrates geographic textual prompts into the inference process, thereby improving the interpretability and generalization of semantic segmentation. SegEarth-OV (Li et al., 2025) introduced a training-free method that learns an efficient feature upsampler from CLIP representations to segmentation masks, significantly reducing both computational and annotation costs. Collectively, these works highlight the potential of OVSS as a flexible and annotation-efficient framework for RS image understanding. However, transferring OVSS from natural images to RS imagery remains challenging due to the domain shifts in spectral characteristics, texture complexity, and object scale. In particular, issues such as domain adaptation, label ambiguity, and fine-grained class representation remain largely unresolved.

### 2.2 *Referring Expression Segmentation (RES)*

RES aims to segment specific objects or regions in an image based on free-form natural language descriptions (e.g., "the small building on the left with a red roof"). Unlike OVSS that focuses on class names, RES targets arbitrary linguistic expressions, often requiring the model to parse fine-grained semantics, resolve spatial relations, and handle diverse object appearances under complex visual contexts. Existing RES methods can be broadly categorized based on their level of supervision. Early works were primarily supervised, relying on large-scale paired image–text annotations. Hu et al. (2016) first introduced the RES task to overcome limitations of conventional semantic segmentation in handling nuanced referential queries. Liu et al. (2017) introduced a recurrent multimodal interaction mechanism to encode the sequential interactions between individual semantic, visual, and spatial information. Hu et al. (2020) proposed a bidirectional cross-modal attention framework, allowing richer interactions between textual and visual features. However, fully supervised RES models would require extensive annotated data for training and inference, which is costly and often domain-specific, leading to poor generalization in out-of-domain scenarios. To address these issues, recent works have explored the training-free approaches. Yu et al. (2023) proposed the first training-free RES framework, enabling zero-shot segmentation from text prompts. Ni et al. (2023) demonstrated that cross-attention maps from a text-to-image generative model can be used to generate segmentation masks, achieving promising performance. Wang et al. (2025) utilized Grad-CAM from a pretrained VLM to produce saliency maps, and introduced an iterative refinement strategy to improve localization accuracy through self-correction.

In the RS domain, RES is still in its infancy, but several recent efforts have laid the foundation for further progress. Yuan et al. (2024) introduced the first large-scale benchmark dataset (RRSIS) for RES task in RS imagery, along with a baseline model that integrates both shallow and deep features to enhance multi-scale context modeling. RMSIN (Liu et al., 2024a) proposed architecture with intra- and cross-scale interaction modules to integrate fine-grained and hierarchical features effectively. RSRefSeg (Chen et al., 2025) transformed semantics from referring expressions to guide the Segment Anything Model (SAM) in producing refined segmentation masks. These efforts demonstrate that RES has strong potential for fine-grained object understanding in RS imagery. However, challenges remain in bridging the modality and domain gaps between natural and remotely sensed images. Although training-free methods have been extensively explored for natural images, similar attempts for RS remain largely absent. This poses a research gap: while prior studies have examined how well VLMs transfer semantic understanding to remote sensing tasks, it remains unclear whether they can effectively identify global spatial context. Addressing this gap is essential to fully harness the power of VLMs for spatially aware, language-guided interpretation in RS domain.

# 3. Method

## 3.1 *Overall framework*

The proposed DGL-RSIS framework provides a unified solution for OVSS and RES according to their specific semantic and contextual requirements. As shown in Fig. 1, DGL-RSIS has three modules: Global-Local Decoupling (GLD), Local Visual-Textual Alignment (LVTA), and Global Visual-Textual Alignment (GVTA). Specifically, the GLD module employs a textual decoupling mechanism that disentangles free-form textual inputs into local semantic tokens and global contextual tokens. Simultaneously, the visual input is decoupled into class-agnostic mask proposals by an unsupervised mask proposal network. The LVTA module performs context-aware feature extraction to obtain discriminative visual patches corresponding to each mask proposal. In parallel, a knowledge-guided prompt engineering strategy enhances the local semantic tokens with domain-specific knowledge. By matching these enriched visual and textual representations, the model assigns semantic labels to the class-agnostic mask proposals, thereby supporting the OVSS task from a local perspective. The GVTA module develops a global-enhanced Grad-CAM to strengthen the global contextual tokens for localizing referring expression. A subsequent mask selection process integrates the pixel-level Grad-CAM activations with mask proposals, ensuring that the selected regions satisfy both the local class semantics (OVSS) and global contextual consistency (RES). Through this hierarchical design, DGL-RSIS forms a unified, training-free framework for universal RS image segmentation.

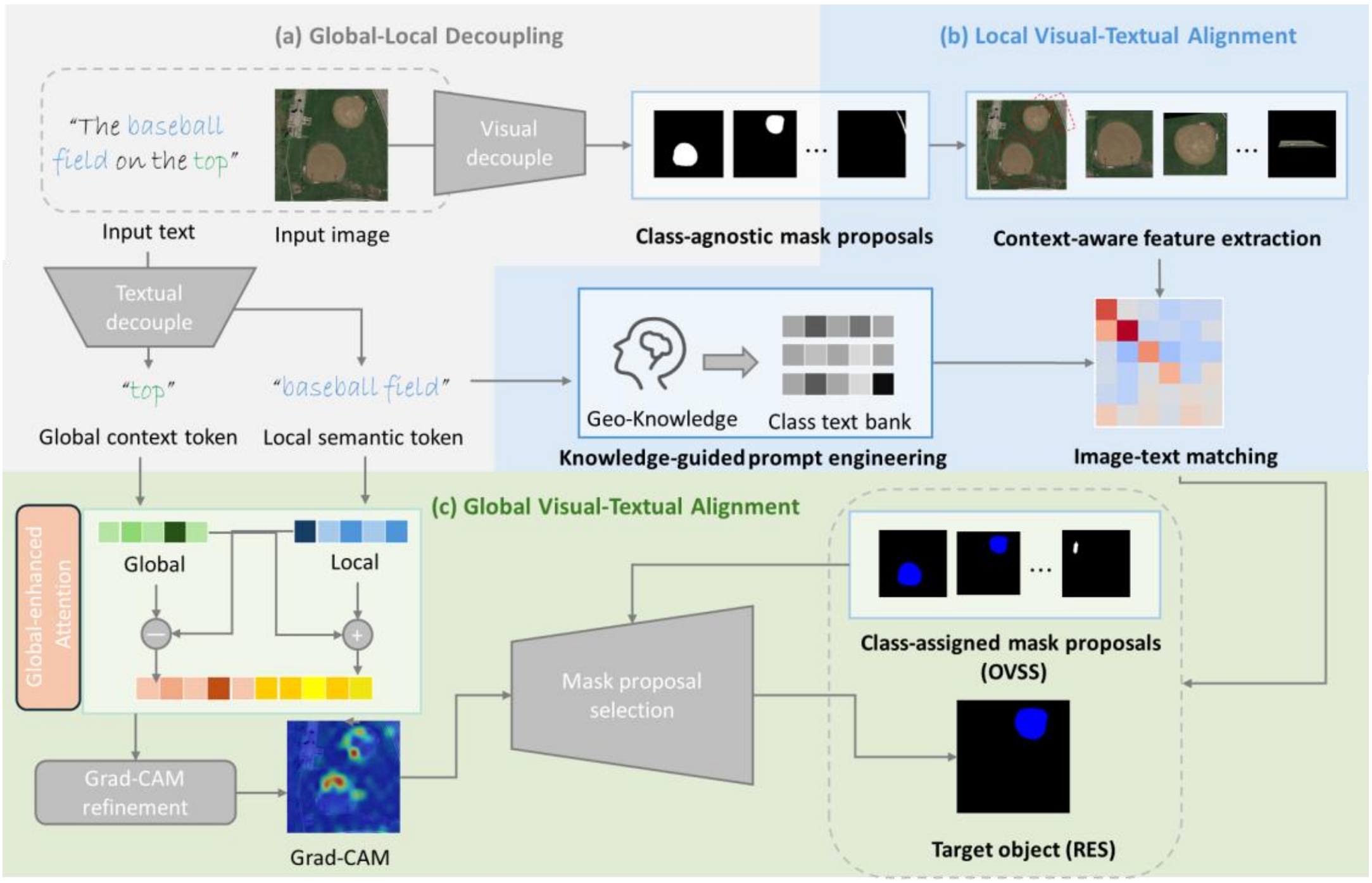

**Fig. 1.** Overall pipeline of the proposed DGL-RSIS framework, consisting of (a) Global-Local Decoupling, (b) Local Visual-Textual Alignment, and (c) Global Visual-Textual Alignment.

## 3.2 *Global-Local Decoupling (GLD)*

To effectively capture both local and global semantics in RS image segmentation, we design a Global-Local Decoupling (GLD) module that simultaneously separates textual and visual inputs into global and local representations. Specifically, textual decoupling and visual decoupling mechanisms are performed in parallel to independently process language and image modalities. This parallel decoupling forms the foundation for subsequent visual–textual alignment at both the global and local levels.

### *3.2.1 Textual decoupling*

Textual inputs vary depending on specific segmentation tasks, which makes it challenging to handle multiple tasks within a unified framework. Generally, a complex textual input $T$ can be divided into class words (e.g., "dock"), which denote object categories, and modifier words, which describe object attributes or spatial relations (e.g., "small"). Inspired by this observation, the proposed textual decoupling mechanism employs a natural language parsing tool, following Liu and Li (2025), to perform part-of-speech tagging and syntactic parsing on the input text and extracts key tokens, including nouns, adjectives, verbs, proper nouns, and numerals. There key tokens can be divided into two groups: local semantic tokens $T_{local}$ and global contextual tokens $T_{global}$. By decoupling the textual inputs into local and global tokens, our framework carries out visual-textual alignment separately, accommodating diverse textual inputs and forming the foundation for universal RS image segmentation.

#### *3.2.2 Visual Decoupling*

Following the textual decoupling mechanism, it is crucial to design differentiated visual processing strategies to address the scale discrepancy between local semantic tokens and global contextual tokens. However, CLIP is inherently designed for image-level representation learning and not directly suitable for pixel-level dense prediction tasks. To bridge this gap, we introduce a visual decoupling mechanism that partitions the image into hierarchical semantic levels.

Specifically, given an input image $I \in \mathbb{R}^{3\times H\times W}$, we employ an unsupervised mask proposal network to generate a set of class-agnostic mask proposals $\mathcal{M}_{ca} \in \mathbb{R}^{M\times H\times W}$, where $M$ is the number of generated mask proposals. These mask proposals enable the image to be decoupled into two semantic scales: 1) global scale, referring to the entire input image $I$ that retains holistic spatial and contextual cues; 2) local scale, referring to individual mask proposals from $\mathcal{M}_{ca}$, where each typically representing a distinct and meaningful geographic entity, enabling localized semantic alignment.

### 3.3 *Local Visual-Textual Alignment (LVTA)*

Building on the GLD module, we reformulate pixel-level dense prediction as mask-level matching. For semantic tasks such as OVSS, the goal is to align category-specific textual semantics with corresponding visual objects. However, transferring VLMs trained on natural images to RS data raises both semantic and visual domain gaps, caused by different object meanings, scales, and top-down perspectives. To address this, we design a knowledge-guided prompt engineering to enhance textual semantics with RS-specific knowledge, and a context-aware feature extraction to adaptively refine visual features. By computing similarity through CLIP, they strengthen local visual–textual alignment for accurate open-vocabulary RS segmentation.

#### *3.3.1 Knowledge-guided prompt engineering*

Prompt engineering has become essential for guiding VLMs in downstream tasks by tailoring textual inputs for semantic alignment. A straightforward strategy is to use predefined class names $T_{org} = \{t_{org}^{j}\}_{j=1}^{C}$ as textual prompts, where $C$ is the number of categories. However, this approach remains suboptimal: 1) class names are often overly ambiguous and abstract to capture the full semantics of RS objects, and 2) most VLMs are pretrained on side-view natural imagery, leading to mismatched textual-visual associations in top-down RS observations.

To mitigate these issues, we introduce a knowledge-guided prompt engineering mechanism that infuses domain-specific knowledge from RS image interpretation into the textual space, forming a comprehensive class text bank $T_{ctb}$. This text bank integrates multiple knowledge sources to refine the semantic representation of prompts:

1) synonym set $\mathrm{T_{syn}} = \{\mathrm{t_{syn}^{j}}\}_{\mathrm{j=1}}^{\mathrm{C}}$: expands ambiguous class labels with semantically related synonyms (e.g., adding "car" for "small vehicle"), reducing lexical uncertainty and enhancing recognition robustness;

2) visual interpretation descriptions $\mathrm{T_{desc}} = \{\mathrm{t_{desc}^{j}}\}_{\mathrm{j=1}}^{\mathrm{C}}$: incorporates concise yet informative textual descriptions distilled from expert RS interpretation (e.g., "tennis court: small bright rectangular area with clear boundaries"), improving fine-grained semantic precision.

3) background land cover classes $\mathrm{T_{bg}} = \{\mathrm{t_{bg}^{j}}\}_{\mathrm{j=1}}^{\mathrm{C}}$: introduces background-related land cover terms to suppress spurious activations and reinforce boundary discrimination.

By embedding this multi-source knowledge into textual embeddings, the proposed strategy effectively bridges the semantic gap between natural and RS domains. The enriched prompts substantially enhance visual–textual similarity matching and enable stronger open-vocabulary generalization across diverse RS segmentation scenarios.

*3.3.2 Context-aware feature extraction*

Compared with objects in natural imagery, RS objects pose unique challenges: 1) they are often small and exhibit substantial scale variations, and 2) they are captured from a top-down perspective, where surrounding spatial context plays a crucial role in category interpretation.

To tackle these issues, we design a context-aware feature extraction strategy that adaptively balances local focus and contextual awareness. Given an input image $I$ and its class-agnostic mask proposals $\mathcal{M}_{ca}$, we generate localized sub-images $I_{local} \in \mathbb{R}^{M\times3\times H\times W}$ by cropping each mask's minimum bounding rectangle (MBR) via $\Phi_{MBR}$ and enlarging it through a controlled scaling transformation $\Phi_{scale}$:

$$I_{local} = \Phi_{scale}(\Phi_{MBR}(I \odot \mathcal{M}_{ca}); \lambda), \qquad (1)$$

where $\lambda > 1$ is the scaling factor.

Unlike conventional axis-aligned bounding box (BB) cropping, the MBR provides a tighter geometric fit for irregular RS objects, while the scaling operation expands the region to retain surrounding spatial cues. This design preserves critical contextual information such as neighboring land-cover transitions, thereby enhancing the encoder's ability to extract discriminative and context-rich local representations. As illustrated in Fig. 2, the proposed strategy achieves a better trade-off between background suppression and contextual preservation, significantly improving feature quality for subsequent alignment.

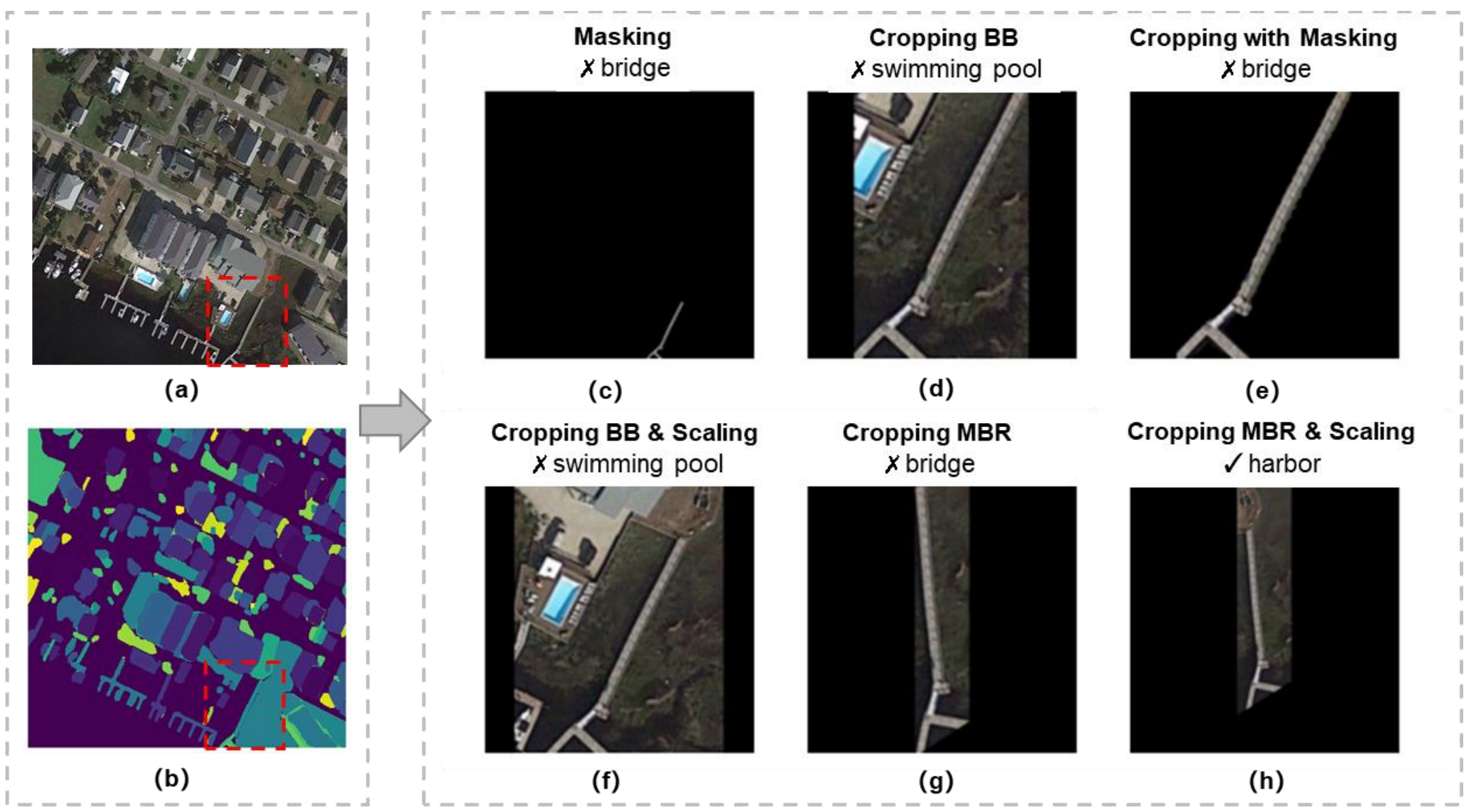

**Fig. 2**. Comparison between conventional cropping strategies and the proposed context-aware feature extraction method. (a) RGB image. (b) Class-agnostic masks. The conventional strategies include: (c) Masking, (d) Cropping by bounding box (BB), and (e) Cropping BB with mask refinement. The context-aware feature extraction strategies include: (f) Cropping BB with scaling transformation, (g) Cropping minimum bounding rectangle (MBR), and (h) Cropping MBR with scaling transformation (ours).

3.4 *Global Visual-Textual Alignment (GVTA)*

In GVTA, the class-agnostic mask proposals are assigned with local semantic labels, enabling open-vocabulary semantic segmentation (OVSS). While OVSS focuses on local category-level alignment, referring expression segmentation (RES) requires both local discrimination and global contextual reasoning. To address this challenge, we propose a GVTA module that introduces a global-enhanced Grad-CAM to explicitly identify image regions corresponding to global contextual tokens. Through the subsequent mask selection module, pixel-level Grad-CAM activations are integrated into mask-level segmentation outputs, achieving accurate and interpretable alignment across both global and local dimensions.

*3.4.1 Global-Enhanced Grad-CAM*

Interpretability techniques, particularly Grad-CAM, have proven effective for interpreting the reasoning process of VLMs. For each token $t_k$ in the textual input, the Grad-CAM activation map $\hat{L}^{(k)}$ is computed as:

$$\hat{L}^{(k)} = A^{(k)} \odot G^{(k)}, \qquad (2)$$

$$G^{(k)} = clamp(\frac{\partial y}{\partial A^{(k)}}, 0, \infty), \qquad (3)$$

where $A^{(k)}$ is the attention map of token $t_k$, $G^{(k)}$ denotes its positive gradients, and $\odot$ is element-wise multiplication. The sentence-level Grad-CAM map is obtained by averaging all token-level maps:

$$\hat{L}^{T} = \frac{1}{|T|} \sum_{k=1}^{|T|} \hat{L}^{(k)}. \qquad (4)$$

Conventional Grad-CAMs typically consider the entire textual input as a single unit, which can yield suboptimal interpretations for RES tasks where modifier words often convey critical spatial and relational cues for disambiguation. As mentioned in Section 3.2.1., the textual input is decoupled into local semantic tokens $T_{local}$ and global contextual tokens $T_{global}$.

Accordingly, we compute the Grad-CAM maps using the VLM's attention maps $A$ and corresponding gradients $G$:

$$\hat{L}^{T_{global}} = A^{T_{global}} \odot G^{T_{global}}, \qquad (5)$$

$$\hat{L}^{T_{local}} = A^{T_{local}} \odot G^{T_{local}}, \qquad (6)$$

where $\hat{L}^{T_{global}}$ and $\hat{L}^{T_{local}}$ highlight regions activated by global and local components, respectively (as shown in Fig. 3).

To strengthen the global contextual reasoning, we compute the L2-normalized activation difference between global and local attention maps:

$$A_{dif} = \frac{A^{T_{mod}} - A^{T_{cls}}}{\left\| A^{T_{mod}} - A^{T_{cls}} \right\|_2}. \qquad (7)$$

This activation difference map $A_{dif}$ captures regions uniquely emphasized by global modifiers. The enhanced global Grad-CAM is then formulated as:

$$\hat{L}^{T}_{enh} = A_{dif} \odot G^{T_{mod}} \odot \hat{L}^{T_{mod}}. \qquad (8)$$

This enhanced Grad-CAM focuses more strongly on global context cues, such as spatial positions and relational descriptions, thereby improving the model's capacity for fine-grained object localization under complex referring expressions. To further refine the segmentation, we adopt a Grad-CAM refinement strategy inspired by Wang et al.(Wang et al., 2025) to transfer pixel-level activations into mask-level segmentation outputs.

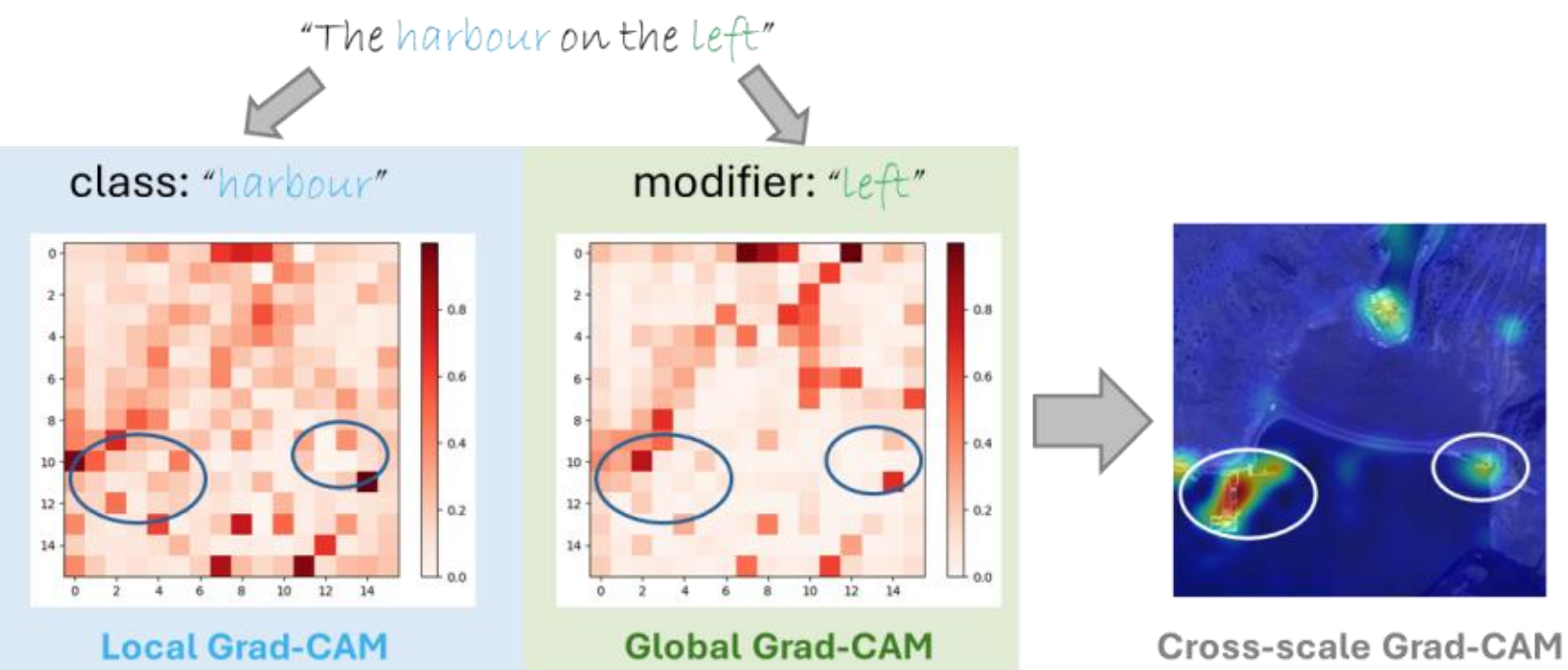

**Fig. 3.** Principle of global-enhanced Grad-CAM. The local Grad-CAM from $T_{local}$ captures category semantics, while the global Grad-CAM from $T_{global}$ models spatial relationships. Their combination forms a global-enhanced Grad-CAM map that integrates both global and local information.

### 3.4.2 Mask Proposal Selection

Based on the global-enhanced Grad-CAM heatmaps, we identify the global mask proposal $\mathcal{M}_{global}$, which highlights the spatial regions corresponding to the referring expression. For each class-agnostic mask proposal $m_i \in \mathcal{M}_{ca}$, the similarity between its visual embedding and the textual embedding is computed using the CLIP model. The local mask proposal $\mathcal{M}_{\mathrm{local}}$ is then obtained by selecting the mask with the highest similarity score:

$$\mathcal{M}_{\mathrm{local}} = \arg\max_{\mathrm{m_i} \in \mathcal{M}_{\mathrm{ca}}} \cos\left(\mathrm{f_v}(\mathrm{I_{local}}), \mathrm{f_t}(\mathrm{T_{ctb}})\right), \qquad (9)$$

where $\mathrm{f_v}(\cdot)$ and $f_t(\cdot)$ denote the visual and textual embedding functions from CLIP, respectively. This produces local semantic masks that align with textual categories, enabling OVS.

To further extend the framework for RES, we integrate the global contextual cues from Grad-CAM and the local semantic understanding from CLIP matching by intersecting the two sets of masks:

$$\mathcal{M}_{final} = \mathcal{M}_{local} \cap \mathcal{M}_{global}. \qquad (10)$$

This joint selection process ensures that the final RES result simultaneously satisfies global contextual reasoning and local semantic consistency, thereby improving both the precision and robustness of referring expression segmentation in RS imagery.

# 4. Experiments and Results

## 4.1 *Datasets and Metrics*

We evaluate our method on two benchmark datasets: iSAID (Waqas Zamir et al., 2019) for OVSS and RRSIS-D (Liu et al., 2024a) for RES. The iSAID dataset contains 2,806 high-resolution images and 655,451 object instances across 15 categories. Following common practice, images are cropped into 800×800 patches, yielding 1,411 training and 458 testing samples. As our method is training-free, all evaluations are conducted directly on the test set.

To assess zero-shot segmentation, we divide iSAID into 11 seen and 4 unseen categories (‘roundabout’, ‘soccer ball field’, ‘plane’, ‘harbor’). Supervised models are trained only on seen classes, while evaluations include both seen and unseen categories. The RRSIS-D dataset comprises 17,402 samples across 20 categories, with 12,181 for training and 3,481 for testing. Each image (800×800) contains diverse object scales, densities, and orientations, posing significant challenges for RES. Following the iSAID protocol, we evaluate our training-free method directly on the test set and define 15 seen and 5 unseen categories (‘storage tank’, ‘tennis court’, ‘train station’, ‘vehicle’, ‘windmill’). For both datasets, we report mean Intersection over Union (mIoU) as the primary evaluation metric for segmentation performance.

## 4.2 *Implementation Details*

In the GLD module, we employ FastSAM (Zhao et al., 2023) for iSAID dataset and Mask2Former (Cheng et al., 2022) for RRSIS-D dataset to generate class-agnostic mask proposals. The choice of mask proposal network is determined based on preliminary experiments, where each model’s ability to handle dataset-specific object scales and densities was carefully evaluated. For local visual-textual alignment, we adopt the CLIP model provided by OpenCLIP (Ilharco et al., 2021), using the ViT-H/14 architecture. This model offers high-quality joint visual–textual embeddings, enabling effective zero-shot mapping between image regions and textual tokens without any additional training. All experiments are conducted on a 40 GB NVIDIA A100 GPU.

# 5. Results

## 5.1 *Segmentation Performance*

We conduct OVSS and RES experiments on the iSAID and RRSIS-D dataset, respectively. To qualitatively evaluate the performance of our method, we visualize the OVSS and RES prediction results and compare them with ground truths and the initial class-agnostics mask proposals. As shown in Fig. 4, the mask proposal network successfully generates class-agnostic masks that correspond to geographic entities. However, since these proposals are class-agnostic, some of them do not perfectly align with individual complete instances, and may split a single object into multiple masks (e.g. "harbor" in Fig. 4 (a) and "roundabout" in Fig. 4 (b)). In the local visual-textual alignment module, our method successfully assigns correct semantic categories to these masks and distinguishes nearby instances, achieving strong OVSS results (Fig. 4 (a-b)). Notably, our method identifies small objects accurately such as ships (Fig. 4 (a)) and small vehicles (Fig. 4 (b)), demonstrating its fine-grained segmentation capability. In the following global visual-textual alignment module for RES task, our method further selects target objects from the class-assigned mask proposals based on detailed referring expression. As shown in Fig. 4 (c), our method identifies the target ("baseball field") along with the correct spatial direction ("on the right"). In Fig. 4 (d), even though the text input includes both spatial relationship and another object ("stadium"), our method successfully identifies the correct target ("ground track field"), demonstrating its capable of tackling relatively complex textual inputs.

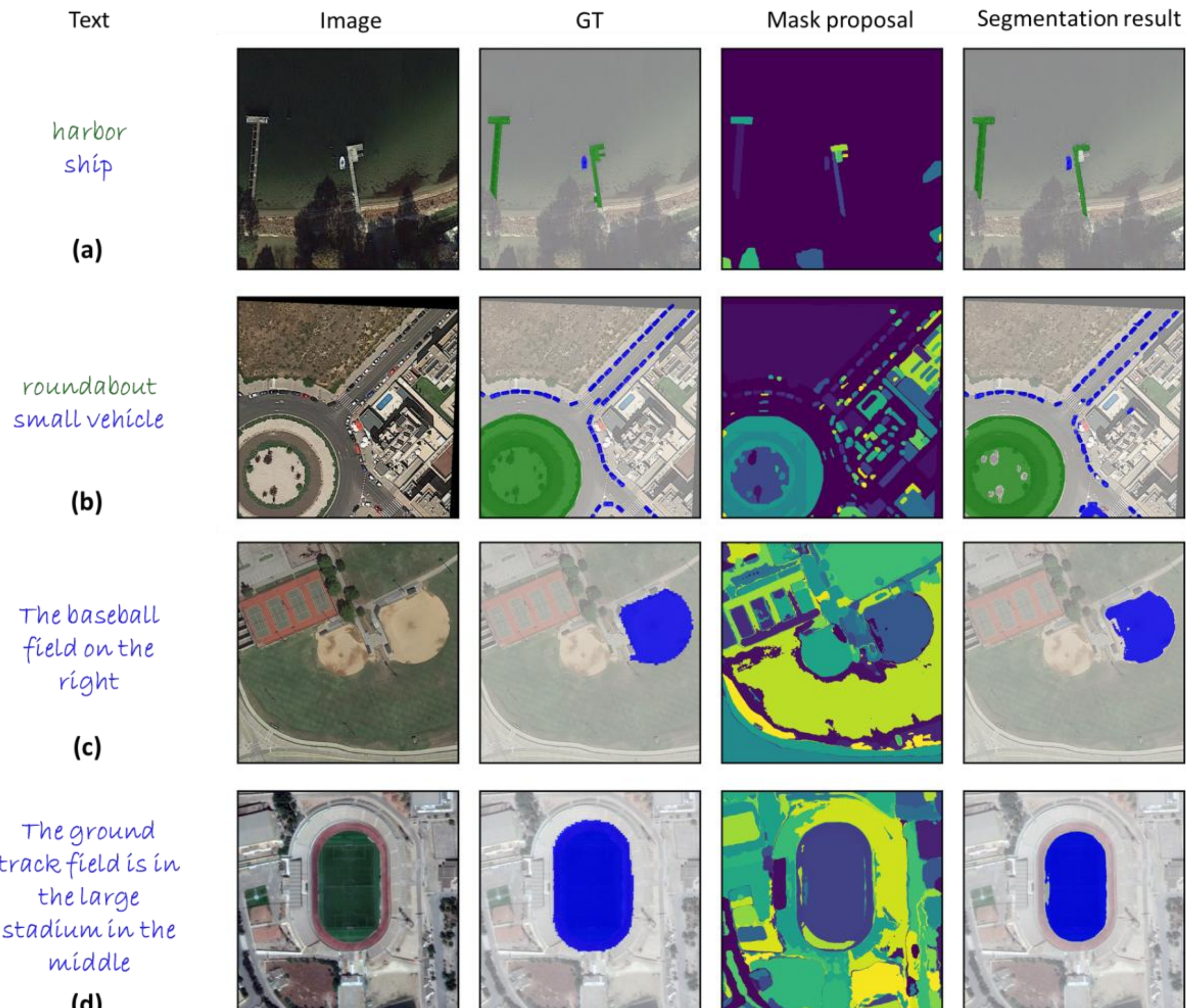

**Fig. 4.** Examples of input text and image, ground truth (GT), class-agnostic mask proposals, and segmentation prediction. (a) and (b) show OVSS examples on the iSAID dataset; (c) and (d) show RES examples on the RRSIS-D dataset.

We report both the overall mIoU and the per-category mIoU on the iSAID and RRSIS-D datasets. As shown in Fig.  (a), for the OVSS task on the iSAID dataset, our method achieves an overall mIoU of 21.55%. Certain categories, such as tennis court and basketball court exhibit relatively high accuracies, reaching 32.10% and 31.48%, respectively. In contrast, six categories perform below average, with bridge being the most challenging class, achieving only 1.26% mIoU. For the RES task on the RRSIS-D dataset (Fig. 5 (b)), our method achieves an overall mIoU of 21.50%, with categories such as storage tank (36.98%), chimney (35.43%) and ship (35.14%) showing relatively high performance. Overall, despite being training-free and not fine-tuned on RS data, our model demonstrates competitive performance across both OVSS and RES tasks.

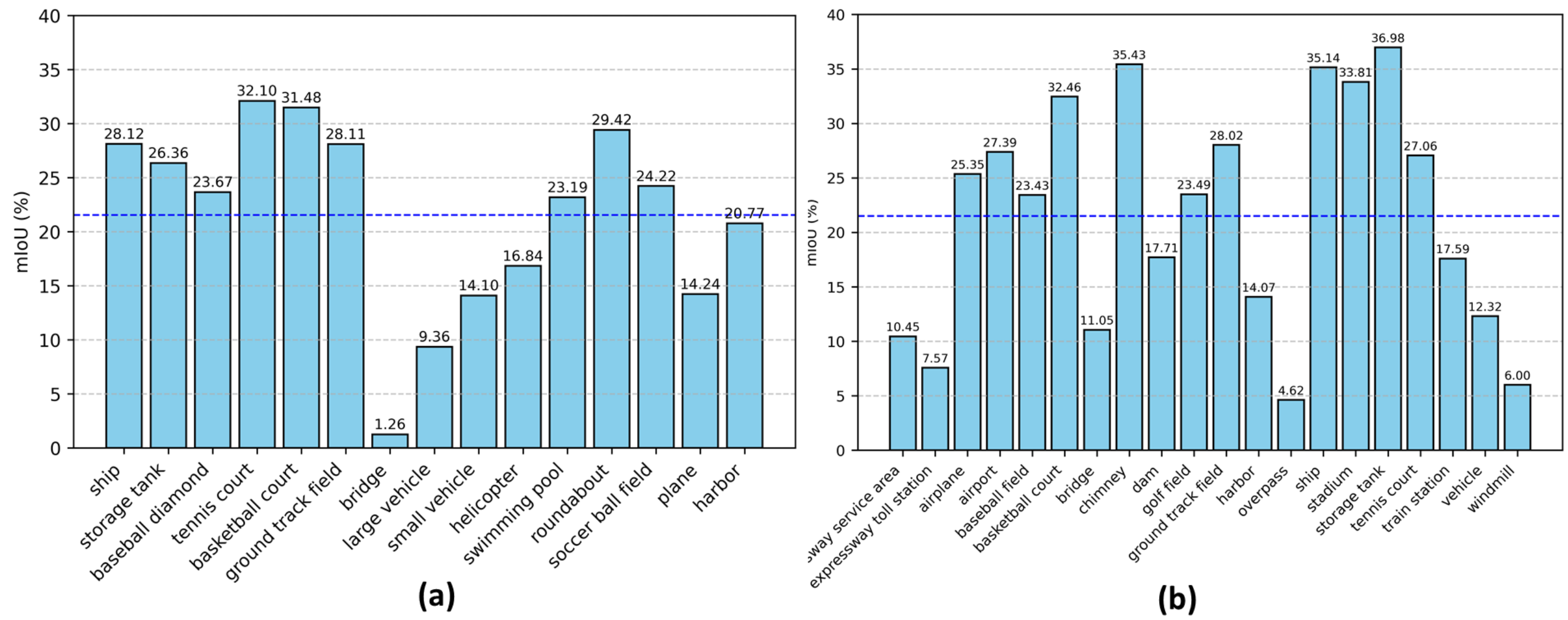

**Fig. 5.** Overall and per-category mIoU on the RS imagery. (a) OVSS results with semantic labels on the iSAID dataset. (b) RES results with semantic labels on the RRSIS-D dataset.

### 5.2 *Comparison with Training-Free and Supervised Methods*

We compared our method with three training-free OVSS approaches, namely SegEarth-OV (Li et al., 2025), MaskCLIP (Zhou et al., 2022), and GEM (Bousselham et al., 2024), as well as three training-free RES approaches, IteRPrimE (Wang et al., 2025), CaR (Sun et al., 2024) and Global-Local CLIP (Yu et al., 2023). These methods represent the state of the art (SOTA) in either remote sensing or natural image segmentation under zero-shot settings. As summarized in Table 1, for the OVSS task on the iSAID dataset, our method achieves the highest mIoU (21.55%) amongst all training-free baselines. The second-best performance is from SegEarth-OV, with an mIoU of 18.40%. Similarly, for the RES task on the RRSIS-D dataset, our method attains an mIoU of 21.50%, outperforming all other training-free methods by a margin of 0.54% to 6.49%. These results demonstrate the effectiveness and robustness of our approach across both OVSS and RES tasks.

**Table 1.** mIoU Comparison between our method and existing training-free and supervised Methods.

| Method | Task | mIoU | Method | Task | mIoU |
|---|---|---|---|---|---|
| Ours | OVS | **21.55%** | Ours | RES | **21.50%** |
| SegEarth-OV | | 18.40% | IteRPrimE | | 20.96% |
| MaskCLIP | | 16.91% | CaR | | 18.39% |
| GEM | | 7.56% | Global-Local CLIP | | 15.01% |

In addition, we also benchmark our method with two supervised OVSS methods, namely, namely ZegFormer (Ding et al., 2022) and CLIPSeg (Lüddecke and Ecker, 2022), as well as two supervised RES methods, RMSIN (Liu et al., 2024b) and RSRefSeg (Chen et al., 2025). As described in Section 5.1.1, these models were trained using only seen categories from the training set, and evaluated on unseen categories, enabling a fair zero-shot comparison. As shown in Fig. 5 (a), for the OVSS task, ZegFormer and CLIPSeg achieve low unseen mIoU scores of 4.06% and 8.82%, respectively, significantly lower than the 21.55% of our method. Moreover, our method exhibits the smallest performance gap between all and unseen categories (0.61%), with both metrics exceeding 20%. In contrast, ZegFormer and CLIPSeg show significantly larger gaps of 14.18% and 15.10%, respectively. For the RES task (Fig. 6 (b)), while RMSIN and RSRefSeg outperforms our method in both overall and unseen categories, they exhibit substantial performance drops on unseen categories, with gaps of 32.76% and 34.30%. In comparison, our method maintains a much smaller gap of 1.51%, indicating greater robustness in zero-shot generalization.

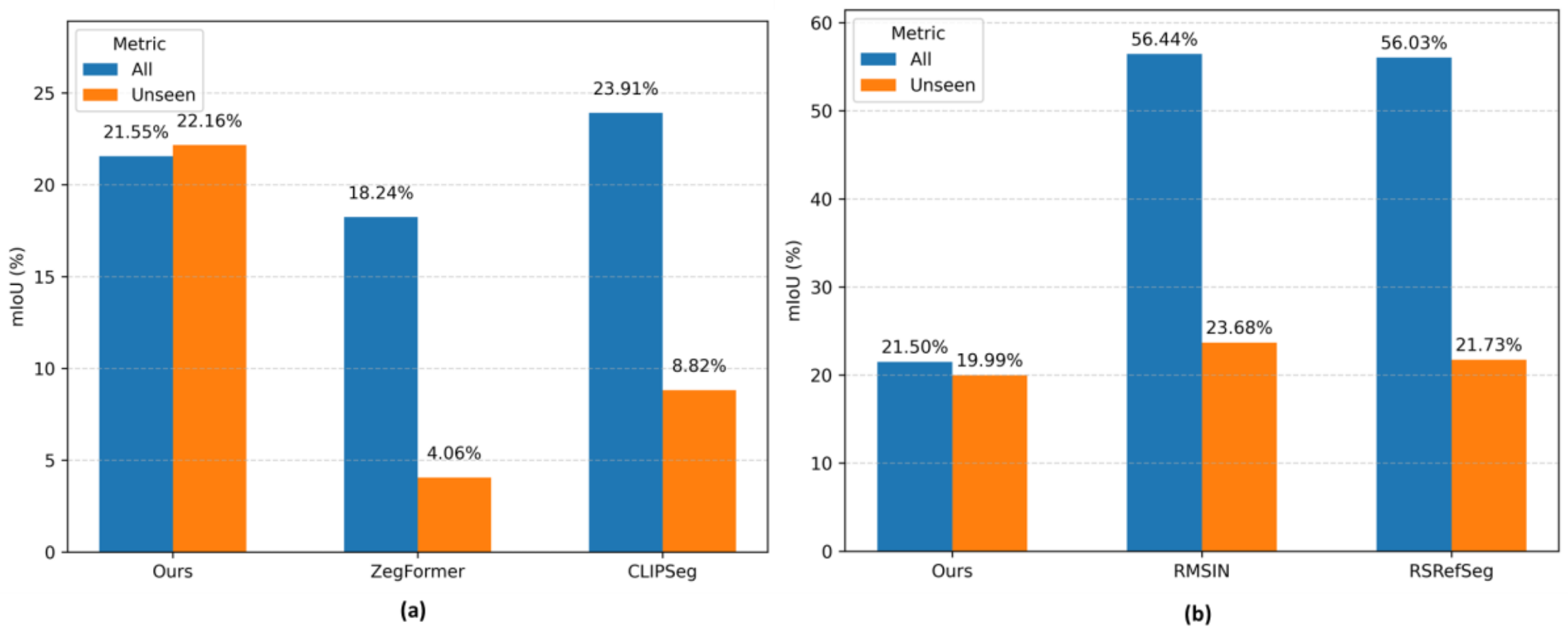

**Fig. 5.** mIoU comparison between all categories and unseen categories for our method and supervised baselines on (a) OVSS task and (b) RES task.

## 5.3 *Ablation Experiments*

### *5.3.1 Knowledge-Guided Prompt Engineering*

As detailed in Section 3.3.1., we introduce a knowledge-guided prompt engineering approach, including the design of a more suitable prompt template for RS and augmentation of input text based on domain knowledge. Ablation studies on the iSAID dataset evaluate the effects of its components. As shown in Table 2, when none of these modules are used, the mIoU drops to 16.78%. Using each module individually improves performance by 0.99%–3.24%, with background classes yielding the largest gain. Combining all three modules results in the best performance of 21.55%, a 4.77% increase over the baseline, demonstrating the effectiveness of our augmentation strategy.

**Table 2.** Ablation Results of prompt engineering methods on the iSAID Dataset.

| Synonyms $T_{syn}$ | Background land cover classes $T_{bg}$ | Visual interpretation description $T_{desc}$ | mIoU |
|---|---|---|---|
| | | | 16.78% |
| √ | | | 18.29% |
| | √ | | 20.02% |
| | | √ | 17.77% |
| √ | √ | √ | **21.55%** |

### *5.3.2 Context-aware Feature Extraction*

As described in Section 3.3.2., we propose a context-aware feature extraction method for the LAVT module. We conduct ablation studies on the iSAID dataset to evaluate the effectiveness of each component in the method and their combinations, and further compare them with commonly-used techniques. As listed in Table 3, using only the mask technique performs poorly on RS images, achieving an mIoU of 6.82%. When combined mask technique with bounding box (BB) cropping, the accuracy increases to 16% but remains 3.22% lower than using BB without mask technique. In contrast, the proposed scaling transformation consistently enhances the model's capability: regardless of the cropping strategy, adding scaling transformation yields a performance gain of 2.01%–2.77%. When combined MBR cropping and buffer expansion scaling transformation, the strategy achieves the highest mIoU of 21.55%, outperforming all other methods and validating the effectiveness of our context-aware cropping framework.

**Table 3.** Ablation results of context-aware feature extraction methods and comparison with other methods.

| Method | | | mIoU |
|---|---|---|---|
| Cropping | Scaling transformation | Masking | |
| — | | √ | 6.82% |
| BB | | √ | 16.00% |
| BB | | | 19.22% |
| BB | √ | | 21.23% |
| MBR | | | 18.78% |
| MBR | √ | | **21.55%** |

* BB: Bounding box; MBR: Minimum bounding rectangle.

#### *5.3.3 Grad-CAM and Mask Proposal Selection*

In Section 3.4., we propose a GVTA module with two key components: global-enhanced Grad-CAM and mask proposal selection. To evaluate their contributions, we conduct ablation studies on the RRSIS-D dataset, as shown in Table 4. We compare our global-enhanced Grad-CAM with baseline Grad-CAM that generates the heatmap using the full text input without focusing on any particular component. Our results show that global-enhanced Grad-CAM improves mIoU by 1.93% and 1.24% over baseline Grad-CAM, with and without the mask proposal selection module, respectively. Additionally, the mask proposal selection module significantly improves segmentation accuracy in both settings: when applied to baseline Grad-CAM, it boosts mIoU from 10.26% to 19.57%, a gain of 9.31%. The best performance (21.50%) is achieved when combining both global-enhanced Grad-CAM and mask proposal selection, validating the effectiveness of the proposed GVAT module.

**Table 4.** Ablation results of the GVAT on the RRSIS-D dataset.

| Grad-CAM | Mask proposal selection | mIoU |
|---|---|---|
| Baseline | | 10.26% |
| Baseline | √ | 19.57% |
| Global-enhanced | | 11.50% |
| Global-enhanced | √ | **21.50%** |

## 6. Discussion

We propose DGL-RSIS, a training-free framework for transferring VLMs trained on natural images to the domain of RS image segmentation. Our approach decouples both image and text modalities into two complementary levels: global contextual and local semantic information. This enables effective adaptation to RS tasks without additional fine-tuning. Prior training-free studies have investigated the transferability of VLMs to RS tasks particularly in terms of semantic class understanding (Li et al., 2025), while overlooking the importance of global context, which is critical for RES. Integrating OVSS and RES within a unified framework is a highly challenging task, primarily due to their different semantic requirements. As discussed in Section 2, OVSS emphasizes fine-grained local class semantics, while RES not only relies on local understanding but also demands comprehension of global spatial relationships. Moreover, as a training-free approach, adapting VLMs pre-trained on natural images to the domain of RS compounds the challenge significantly. To address this gap, DGL-RSIS explicitly integrates both global spatial context and local class semantics into a unified system. To the best of our knowledge, this is the first attempt to bridge semantic and spatial representation learning in a zero-shot setting for RS segmentation. Addressing this missing link is essential to fully leverage the potential of VLMs for spatially-aware, language-guided interpretation in the RS domain.

To investigate effective strategies for transferring a VLM fine-tuned on natural images to RS segmentation tasks, we introduce domain knowledge as an essential component. Knowledge-driven and data-driven models represent two dominant paradigms for RS information extraction (Reichstein et al., 2019). Knowledge-driven approaches rely on prior knowledge from experts or geospatial data and extract information through reasoning, but are often limited by the scarcity of such knowledge relative to the complexity of natural phenomena (Yuan et al., 2020). In contrast, data-driven approaches, especially deep learning, extract information from large datasets but depend heavily on data quality and often struggle with generalization (B. Li et al., 2024). Recognizing the respective strengths and limitations of these two paradigms, the integration of domain knowledge with data-driven deep models has received increasing attention. For example, Ge et al. (2022) propose a paradigm of geoscience-aware deep learning aiming to leverage domain-specific knowledge to enhance the performance of deep models for RS information extraction. However, existing efforts have been largely concentrated in the visual domain. The incorporation of domain knowledge into VLMs remains rare. Our work addresses this gap by investigating how domain knowledge can be explicitly injected into VLM-based segmentation pipelines, thereby improving model performance and interpretability in RS applications.

At the local visual-textual alignment level, we emphasize the incorporation of domain knowledge. For the visual inputs, we observe that standard cropping strategies commonly used for natural images are suboptimal when applied to remote sensing imagery (Fig. 2). To address this, we propose a context-aware feature extraction method, considering that object recognition in RS imagery requires supporting contextual cues. Based on this insight, we introduce a scaling transformation strategy, which yields performance gains of 2.01%–2.77% regardless of the cropping method used (Table 3). This supports our intuition that contextual information is particularly beneficial for object recognition in RS scenarios. On the textual inputs, we propose knowledge-guided prompt engineering. Inspired by the knowledge of remote sensing imagery, we design a new prompt template and compare it with commonly used prompts for both natural and remote sensing images (Table 2). The improved performance from our proposed prompt highlights the need to translate RS semantics into forms that VLMs can better comprehend. Therefore, we further introduce a RS knowledge-guided text augmentation strategy. To better illustrate the effectiveness of our text augmentation strategy, we present CLIP similarity heatmaps in Fig. 6. In Fig. 6 (a), using the baseline text input, the model misclassifies “harbor” as “background” and confuses “tree” with “small vehicle”. In Fig. 6 (b), after applying the synonyms (e.g., replacing “harbor” with “dock”), the similarity improves and leads to correct classification. In Fig. 6 (c), incorporating background land cover classes enriches background descriptions and enhances the identification of non-target categories such as “tree”. Finally, Fig. 6 (d) shows that incorporating visual interpretation descriptions further clarifies category distinctions, as reflected in a more accurate similarity matrix.

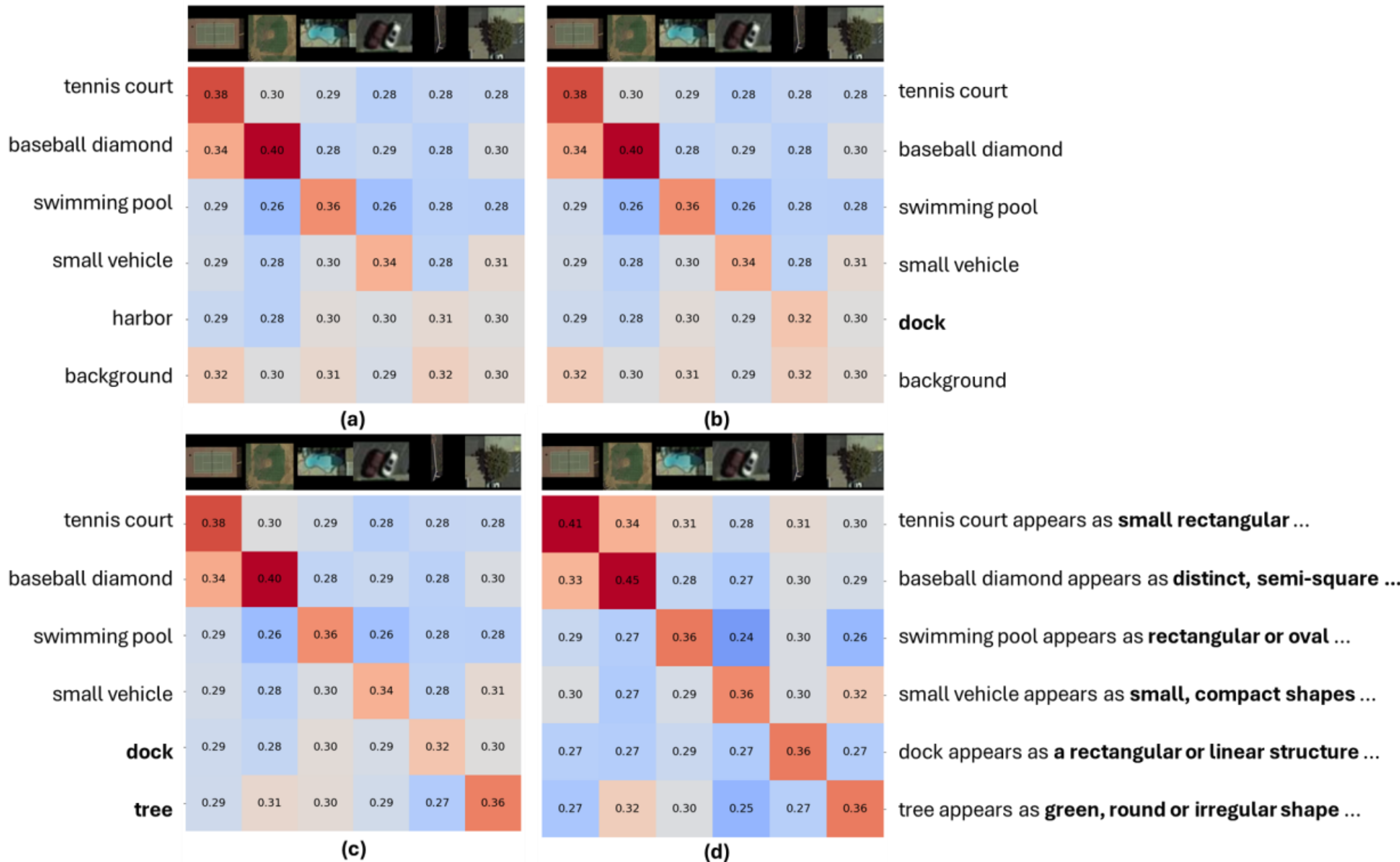

**Fig. 6.** CLIP similarity heatmaps for different category text inputs. (a) Original class names; (b) Original names enhanced with synonyms; (c) Further addition of background land cover classes based on (b); (d) Replacement of class names with visual interpretation descriptions on top of (c).

At the global visual-textual alignment level, we explicitly identify image regions corresponding to global modifiers in the referring expression. As mentioned in Section 3.2, conventional Grad-CAM treats the entire input text equally, assigning uniform attention to each word. However, since global modifier words play a crucial role in RES, we enhance their impact by proposing global-enhanced Grad-CAM module. Ablation results in Table 4 demonstrate the effectiveness of this enhancement. To further analyze the underlying reasons, we visualize the results of baseline and global-enhanced Grad-CAM in Fig. 7. The global-enhanced version, which emphasizes global modifiers, produces more focused and accurate heatmaps aligned with the intended target regions. In contrast, the baseline Grad-CAM, treating all words equally, suffers from attention dispersion and misidentification.

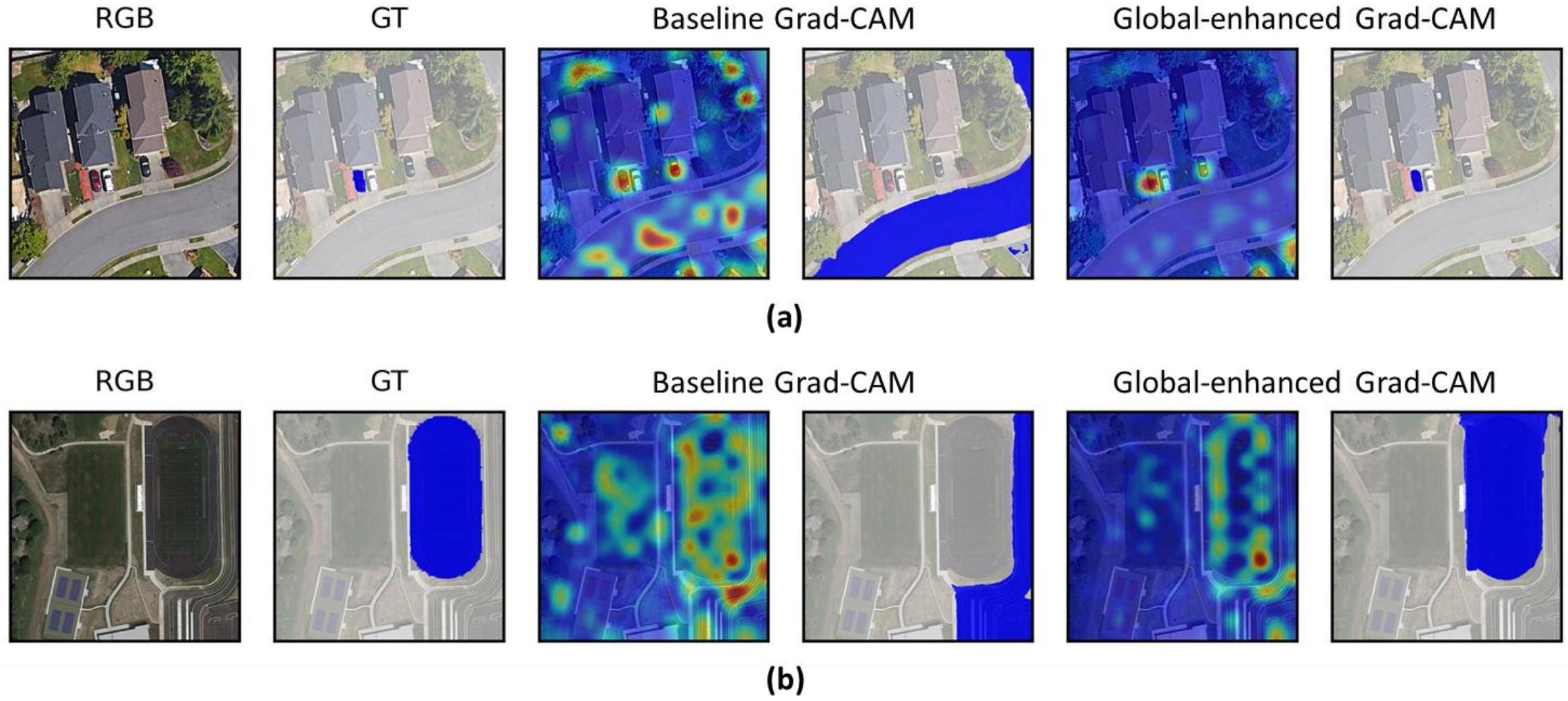

**Fig. 7.** Visual comparison between baseline Grad-CAM and global-enhanced Grad-CAM. In (a), the input text is "A red vehicle"; in (b), the input text is "An oval ground track field".

Our method demonstrates strong potential for practical application, especially in scenarios requiring the adaptation of VLMs from the natural image domain to remote sensing. By explicitly addressing the domain gap through the GLD module, we offer a feasible and effective solution for vision-language transfer. The framework is flexible, modular, and annotation-free, making it suitable for large-scale or resource-constrained RS tasks. The decoupled design for local and global visual-textual alignment could inspire future research on interpretable, cross-domain vision-language architectures. Thanks to its training-free nature and incorporation of domain knowledge, our framework can be further extended to specialized applications. For example, it can be adapted for agricultural mapping by integrating farming-related knowledge, or for habitat mapping with ecological context-aware prompts and cues.

While our training-free method achieves acceptable accuracy, there is still a noticeable gap compared to fully supervised approaches. Future research could explore incorporating supervision into this framework to enhance performance, while still preserving the decoupled local-global alignment design. Such integration could help mitigate the overfitting or bias issues toward seen categories often observed in traditional supervised methods, and further improve the generalization capability on RS image segmentation tasks. Beyond segmentation, we see this work as a general paradigm shift in how VLMs can be adapted to RS tasks. Its flexible design makes it applicable to a wide range of applications, including classification, detection, and pattern recognition, as well as broader geospatial reasoning and decision-making scenarios.

# 7. Conclusion

In this paper, we propose DGL-RSIS, a training-free framework for universal remote sensing image segmentation, capable of handling both open-vocabulary semantic segmentation (OVSS) and referring expression segmentation (RES). The framework decouples visual and textual inputs into local semantic tokens and global contextual tokens via the Global-Local Decoupling (GLD) module, while generating class-agnostic mask proposals through an unsupervised network. For local visual-textual alignment, we employ context-aware feature extraction and knowledge-guided prompt engineering to match enriched textual features with mask-guided image patches, enabling zero-shot classification for OVSS. For global visual-textual alignment, a global-enhanced Grad-CAM module highlights image regions corresponding to global contextual tokens, and a mask selection process integrates these cues with local mask proposals, ensuring consistency between local semantics and global context for RES. Through this hierarchical design, DGL-RSIS unifies both tasks in a single framework, demonstrating effective zero-shot segmentation across diverse remote sensing datasets.

Extensive experiments demonstrate that our framework achieves strong generalization across both OVSS

and RES benchmarks, outperforming previous SOTA methods and even surpassing some supervised approaches. Ablation studies validate the effectiveness of each module, while interpretability experiments further reveal how local and global alignments contribute to final predictions.

Although our framework is training-free, which limits its absolute accuracy, it establishes a solid foundation for future work. A promising research direction lies in extending this framework to supervised settings, with a particular focus on mitigating bias toward seen categories. Given its satisfactory performance without training, our framework offers a scalable and adaptable paradigm for RS segmentation tasks in both research and practical scenarios.

## Acknowledgement

The authors are grateful for the financial support provided by China Scholarship Council–University of Bristol PhD Scholarships Program.